\documentclass[conference, review]{IEEEtran}
\IEEEoverridecommandlockouts

\usepackage{amsmath,amssymb,amsfonts}
\usepackage{algorithmic}
\usepackage{graphicx}
\usepackage{textcomp}
\usepackage{xcolor}
\usepackage{booktabs}
\usepackage{amssymb}
\usepackage{pifont}
\newcommand{\cmark}{\ding{51}}%
\newcommand{\xmark}{\ding{55}}%
\definecolor{LightCyan}{rgb}{0.95,0.95,0.95}
\usepackage{xurl}

\usepackage{acronym}

\usepackage[detect-all=true]{siunitx}
\DeclareSIUnit\decibelm{dBm}

\def\BibTeX{{\rm B\kern-.05em{\sc i\kern-.025em b}\kern-.08em
    T\kern-.1667em\lower.7ex\hbox{E}\kern-.125emX}}
\begin{document}

\title{Cross-platform Smartphone Positioning at Museums}

\author{

\IEEEauthorblockN{Alessio Ferrato\IEEEauthorrefmark{1}, Fabio Gasparetti\IEEEauthorrefmark{1}, Carla Limongelli\IEEEauthorrefmark{1},\\Stefano Mastandrea\IEEEauthorrefmark{4}, Giuseppe Sansonetti\IEEEauthorrefmark{1}, and 
Joaquín Torres-Sospedra\IEEEauthorrefmark{2}\textsuperscript{,}\IEEEauthorrefmark{3},   }

\IEEEauthorblockA{
\IEEEauthorblockA{\IEEEauthorrefmark{1}\textit{Department of Engineering}, \textit{Roma Tre University}, Via della Vasca Navale 79, 00146, Roma, Italy}
\IEEEauthorblockA{\IEEEauthorrefmark{4}\textit{Department of Educational Science}, \textit{Roma Tre University}, Via del Castro Pretorio 20, 00185, Roma, Italy}
\IEEEauthorrefmark{2}\textit{Departament d'Informàtica}, \textit{Universitat de València}, Campus de Burjassot, 46100 Burjassot, Spain}
\IEEEauthorblockA{\IEEEauthorrefmark{3}\textit{VALGRAI}, \textit{Valencian Graduate School and Research Network of Artificial Intelligence}, València, Spain}
}

\maketitle 

\begin{abstract}
Indoor Positioning Systems (IPSs) hold significant potential for enhancing visitor experiences in cultural heritage institutions. By enabling personalized navigation, efficient artifact organization, and better interaction with exhibits, IPSs can transform the modalities of how individuals engage with museums, galleries and libraries. However, these institutions face several challenges in implementing IPSs, including environmental constraints, technical limits, and limited experimentation. In other contexts, Received Signal Strength (RSS)-based approaches using Bluetooth Low Energy (BLE) and WiFi have emerged as preferred solutions due to their non-invasive nature and minimal infrastructure requirements. Nevertheless, the lack of publicly available RSS datasets that specifically reflect museum environments presents a substantial barrier to developing and evaluating positioning algorithms designed for the intricate spatial characteristics typical of cultural heritage sites. To address this limitation, we present BAR, a novel RSS dataset collected in front of 90 artworks across 13 museum rooms using two different platforms, i.e., Android and iOS. Additionally, we provide an advanced position classification baseline taking advantage of a proximity-based method and $k$-NN algorithms. In our analysis, we discuss the results and offer suggestions for potential research directions.
\end{abstract}

\begin{IEEEkeywords}
RSS, Fingerprinting, Cultural Heritage, Visitor, Localization, Dataset
\end{IEEEkeywords}

\section{Introduction \& Background}

Indoor Positioning Systems (IPSs) have emerged as an essential solution for environments where Global Positioning System (GPS) signals are weak or unavailable. Among various IPS implementations, those based on Received Signal Strength (RSS) measurements from Bluetooth Low Energy (BLE) and WiFi transmitters have gained significant traction due to their cost-effectiveness, scalability, and minimal infrastructure requirements~\cite{ramires24}. These systems provide person or object position by measuring signal strengths from multiple transmitters and employing different techniques such as proximity~\cite{Kuflik11}, trilateration~\cite{Spachos20} or fingerprinting~\cite{mendoza2018long}.

Cultural heritage institutions such as Galleries, Libraries, Archives, and Museums (GLAMs) represent a particularly compelling application domain for IPSs~\cite{augello2021site}. These particular settings face unique challenges to improve visitor engagement, flow, and personalization that an IPS can address effectively. Moreover, cultural heritage spaces typically have strict preservation constraints that limit the application of contextually dissonant technologies. However, RSS-based IPSs are normally non-invasive solutions that respect these environmental constraints while delivering valuable positioning data.

\subsection{RSS-based IPSs in Cultural Heritage}

Several researchers have explored the integration of RSS-based positioning within cultural heritage contexts. \cite{Kuflik11} implemented a proximity-based IPS in a museum using a wireless sensor network operating on the \SI{2.4}{\giga\hertz} band with the 802.15.4 protocol. Their system utilized three components: fixed Beacons placed on Points-of-Interest (POIs) (i.e., artworks), mobile wearable tags carried by visitors, and Gateways. This approach addressed positioning challenges through a complex combination of hardware configuration, algorithms that balanced signal detection with historical data, and filtering that considered the museum's spatial layout. Their system achieved a high accuracy within a range of \SIrange{1.5}{2}{\meter} and covered POIs across 4 exhibition rooms with 26~Beacons.

\cite{Chianese13} implemented a location-based museum guide that uses a Wi-Fi-based IPS to allow artworks to automatically deliver personalized multimedia content to visitors' smartphones when they approach POIs. Their IPS exploit RSS measurements from multiple access points attached to artworks to determine visitor proximity. The system achieves \SIrange{91}{100}{\percent} accuracy by implementing a two-phase algorithm: first, scanning nearby Wi-Fi networks associated with artworks, then determining the closest artwork based on a weighted average of three consecutive RSS measurements. However, their system was tested in a laboratory room with 11 points simulating a museum.

\cite{jimenez2017finding} studied the effectiveness of Ultra-Wide Band (UWB) and BLE for wayfinding in a museum-like setting (i.e., a corridor with 6 artworks on the walls). The authors found that BLE, along with Pedestrian Dead Reckoning, can help smartphone users navigate to POIs with similar accuracy as UWB.

\cite{gutierrez2019limus} explores acoustic, visible light, and BLE technologies to provide Location-Based Services in museums or archaeological sites. They concluded that BLE beacons have robust detection rate at a radius of about \SI{1.5}{\meter}.

\cite{Spachos20} explained how to create a BLE-based location service in a museum, detailing its potential benefits. They tested their concepts in a lab designed to resemble different museum spaces. They used up to four beacons for distance estimation and location accuracy results confirming the BLE potential in such a setting.

\subsection{Existing RSS Datasets in Cultural Heritage}

Despite numerous works in GLAMs IPSs, publicly accessible RSS datasets specifically recorded in cultural heritage settings remain scarce. While university libraries have been well-represented~\cite{mohammadi2017semisupervised,
mendoza2018long,mendoza2019ble}, museums, galleries, and archives remain underrepresented in public repositories.
To our knowledge, only two datasets specifically recorded in real museum environments have been made publicly available.

\cite{bracco2020museum} published RSS data collected from a WiFi-based positioning system deployed at Uruguay's National Museum of Visual Arts. Their implementation covered multiple floors divided into 16 zones defined by artwork locations rather than physical room boundaries. While valuable, particularly for its focus on accessibility for visually impaired visitors, this dataset lacks comprehensive documentation as the authors prioritized system implementation description.

More recently, \cite{girolami2024bluetooth} introduced a dataset recorded in the Monumental Cemetery's museum in Pisa, Italy. This dataset is meant to test how well proximity detection and crowdsensing positioning algorithms work in real settings~\cite{girolami2024crowdsensing}. The study covered 10 artworks placed \SIrange{1.5}{5}{\meter} apart, with each artwork having its own BLE beacon. The authors used a custom app built with React Native on 8 different Android smartphones. They followed 4 unique visitor routes, completing a total of 32 test runs. The dataset includes raw signal strength measurements and notes on when visitors checked in and out at each artwork. The recordings took place during regular museum hours with other visitors.

Previous studies have made important contributions by using realistic settings, but there are still gaps in research on RSS-based IPS in cultural heritage settings. Many of the aforementioned studies employ or assume that there is one transmitter for each artwork, which is not always possible in real settings. Additionally, most research only uses one or a few Android devices, limiting findings across different operating systems. Finally, many studies on cultural heritage do not share the data limiting reproducibility. To address these gaps, we present and analyze \emph{BAR}, a novel dataset recorded in a real cultural heritage setting. Our main contributions are as follows.

\begin{itemize}
    \item \textbf{Expanded artwork coverage:} Our dataset significantly expands the scope of existing collections by encompassing 90 distinct artworks distributed across 13 museum rooms, addressing the substantial quantitative limitations of previous datasets and enabling more robust algorithmic development for complex museum environments.
    
    \item \textbf{Cross-platform measurements:} Unlike prior works that predominantly relied on Android devices, our dataset incorporates RSS measurements from both iOS and Android smartphones supporting robust cross-platform studies.
    
    \item \textbf{Standardized baselines:} We provide position classification baselines that establish performance benchmarks for future algorithmic developments in museum environments, facilitating fair comparisons.
    
\end{itemize}

\section{Setup and Collection Methodology}
\label{sec:collectionMethodology}

We collected our data inside \textit{Palazzo Barberini}, a palace built during the 17\textsuperscript{th} century in Rome, Italy, which hosts part of the \textit{Gallerie Nazionali di Arte Antica}, the main national collection of historical paintings in the city. We recorded everything during opening hours on different days and hours to ensure a real-world conditions.

\begin{figure}[htbp]
    \centering
    \includegraphics[width=0.90\linewidth]{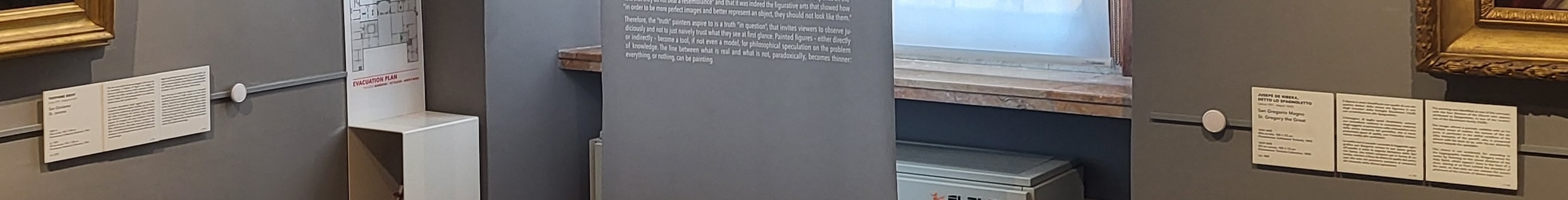}
    \caption{Beacon placement next to artwork labels.}
    \label{fig:position_1}
\end{figure}

\begin{figure}[htbp]
    \centering
    \includegraphics[width=0.90\linewidth]{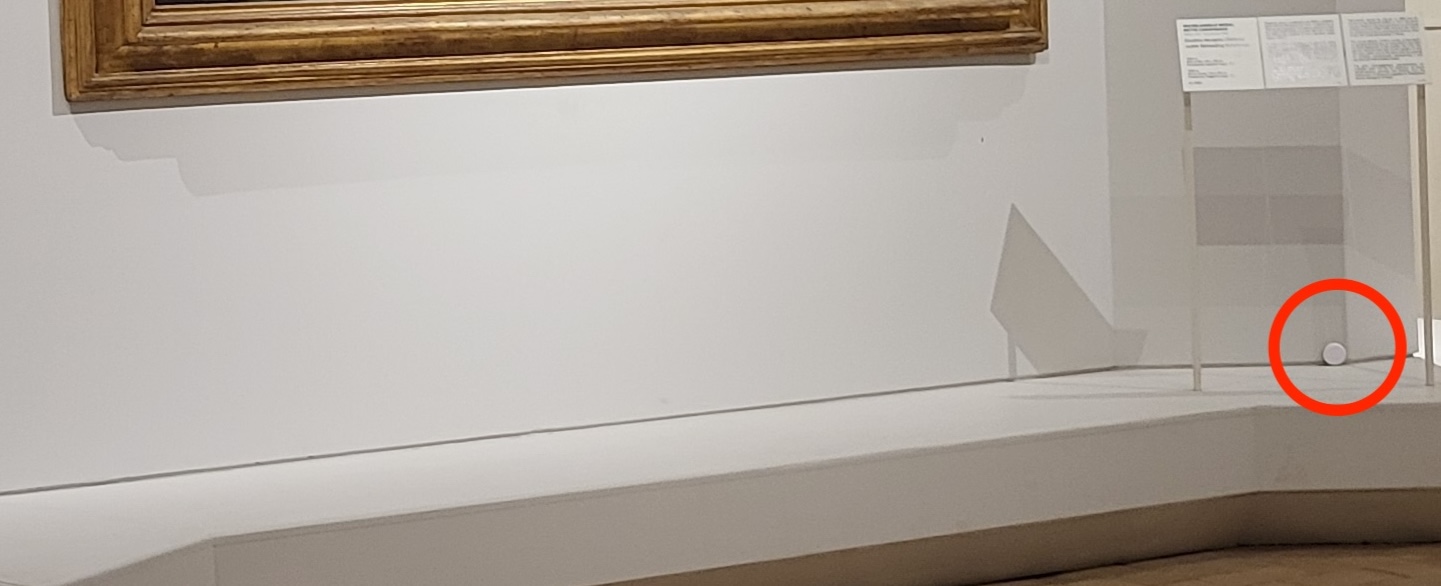}
    \caption{Beacon positioning on platform-mounted artworks. Red circles to highlight the positions.}
    \label{fig:position_2}
\end{figure}

\begin{figure}[h]
    \centering
    \includegraphics[width=0.90\linewidth]{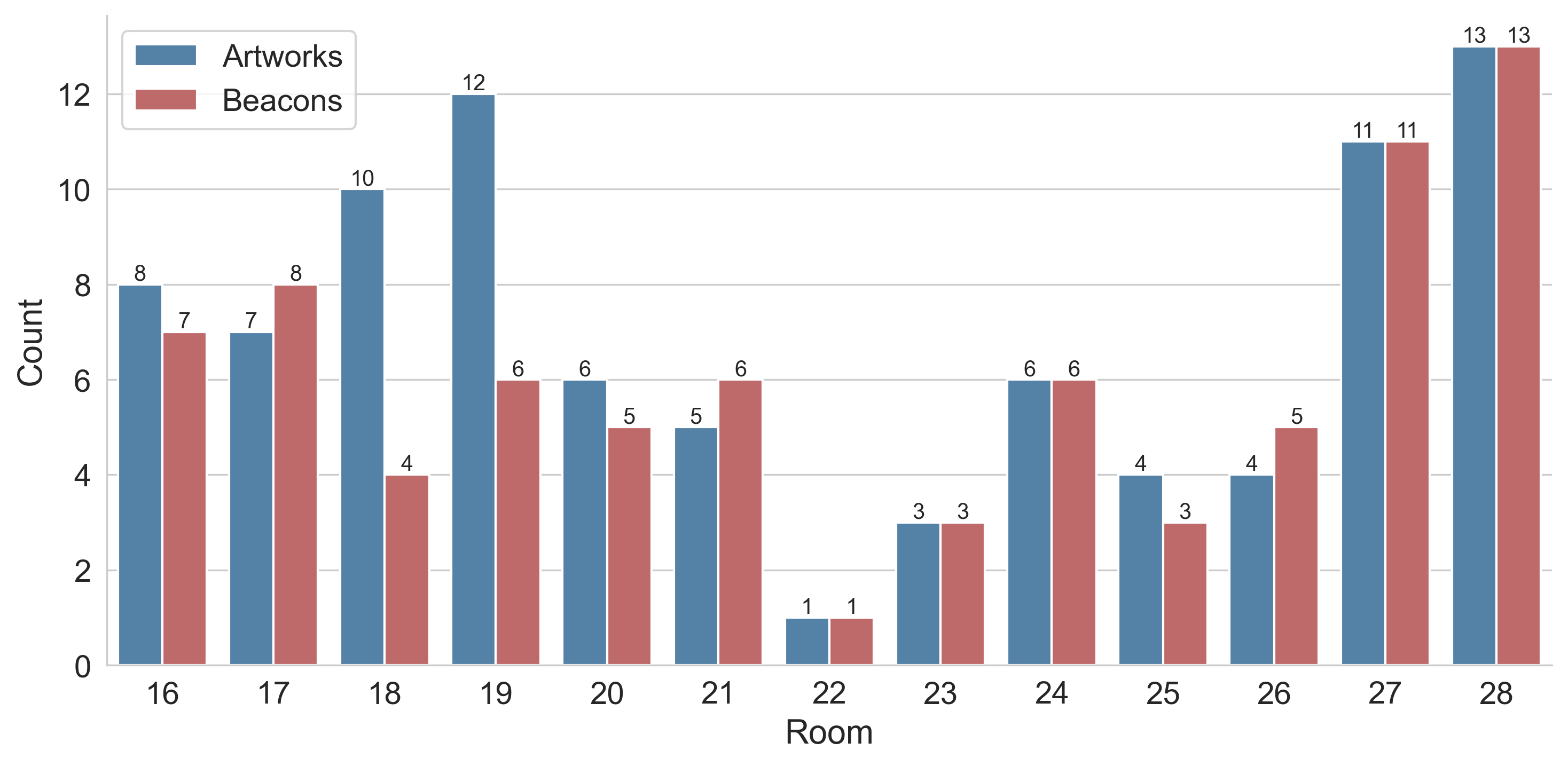}
    \caption{Artwork and beacons distribution for each room.}
    \label{fig:artwork-dist}
\end{figure}

We placed $93$ BLE H2 Beacons made by Moko Smart inside the museum. We have different placement positions to minimize the visual impact of the devices strategically. Most beacons are positioned next to the labels of the artwork (see Fig. \ref{fig:position_1}). Another common placement is on the bottom right corner if the artwork is displayed on a platform (see Fig. \ref{fig:position_2}). Less frequently, beacons are positioned below display cases or in the corners of the rooms. The number of artwork for each room varies from a minimum of 1 to a maximum of 13, Fig.~\ref{fig:artwork-dist} shows the full distribution along with the number of beacons for each room with at least one artwork inside.

Beacons are configured to transmit one iBeacon advertisement every \SI{300}{\milli\second}, utilizing a transmission power of \SI{0}{\decibelm}. Each advertisement is distinctly identified by unique major and minor identifiers, commonly corresponding to the beacon's room and the specific position inside the room while maintaining a consistent UUID across all devices.

For the data collection we developed a cross-platform smartphone application using Flutter. The source code for the application is available on GitHub~\cite{maestro2025flutter}. Unlike other research studies  \cite{torres2014ujiindoorloc,Spachos20,girolami2024bluetooth}, we chose a non-native application to facilitate the development as we need to record data on two different platforms. To capture the iBeacon advertisements, we utilized the \textit{Dchs Flutter Beacon} library. A significant limitation we encountered using this library was the variability in the number of advertisements received across platforms. Specifically, Android devices could capture all available advertisements, while iOS devices were restricted to one advertisement per beacon per second. However, we selected the previous library over the more widely recognized \textit{flutter blue plus} as iBeacons are not supported on iOS. For the collection of smartphone sensor data, we utilized \textit{sensors plus}; although this data is not currently in use, it has been recorded for potential future research applications. The application was installed on two smartphones: an Apple iPhone 16 and a Samsung Galaxy 21 FE Edition.

All recordings were conducted by a single person, who held one phone in each hand. He positioned himself in front of an artwork, selected the corresponding artwork from a list within the application, and started the recording process. During the recording, he changed the distance from the artwork by moving closer and farther away, as well as shifting left and right to capture various viewpoints of the artwork. The recordings took place on different days, with the devices being alternated between hands to mitigate potential bias.

Table~\ref{tab:datasets} summarizes \emph{BAR}, our dataset, with respect to other publicly available datasets recorded in similar settings.

\begin{table}[h] 
\centering 
\caption{Datasets comparison} \label{tab:datasets} \resizebox{\columnwidth}{!}{%
\begin{tabular}{@{}lllcccc@{}} \toprule Dataset & Technology & Platform & Artworks & Transmitters & Rooms & Baseline \\ 
\midrule 
\cite{bracco2020museum} & WiFi & Android & 13 & 15 & \xmark & \xmark \\
\cite{girolami2024bluetooth,girolami2024crowdsensing} & BLE & Android & 10 & 10 & - & \cmark \\ 
\midrule BAR (Our Work) & BLE & Android \& iOS & 90 & 93 & 13 & \cmark \\ 
\bottomrule 
\multicolumn{7}{c}{\cmark stands for present; \xmark stands for not present; - stands for not applicable.}
\end{tabular} } 
\end{table}

\section{Data description}
\label{sec:dataDescription}

\emph{BAR} includes both the raw recordings and the corresponding fingerprints. Since recordings are captured on different days, we have organized the files by platform and session. Each session covers the full artwork collection. Additionally, we have assigned an incremental ID to facilitate the identification (see Table~\ref{tab:subsets}). We will maintain the following format for this version and future iterations:

\begin{itemize}
     \item \textit{\{platform\}\_session\_\{number\}\_\{id\}.json}
     \item \textit{\{platform\}\_session\_\{number\}\_\{id\}.csv}
\end{itemize}
The \textit{json} files contain the unprocessed raw data, whereas the \textit{csv} files contain the processed fingerprints. With this approach, we have $4$ independent subsets of data covering all artworks and rooms. 

Specifically, each \textit{json} file encompasses multiple recordings, each one is related to a specific artwork. The structure for each recording in the file is as follows:
\begin{small}
\begin{verbatim}
{   _id: { $oid: String },
    room: Integer,
    label: String,
    recordingStartTime: String,
    recordingEndTime: String,
    recordingDurationSeconds: Integer,
    platform: String,
    accelerometerData: [{
        x:             Float,
        y:             Float,
        z:             Float,
        timestamp:     String
    }, ...],
    gyroscopeData: [{
        x:             Float,
        y:             Float,
        z:             Float,
        timestamp:     String
    }, ...],
    magnetometerData: [{
        x:             Float,
        y:             Float,
        z:             Float,
        timestamp:     String
    }, ...],
    bleData: {
        "12-1": [{
            rssi:      Integer,
            timestamp: String
        },..],  ...,
    },
}
\end{verbatim}
\end{small}

We extracted RSS fingerprints using a sliding window approach. For each beacon, we define a window of fixed duration \( w \) and a step size \( s \), where the overlap between consecutive windows is given by \( o = w - s \).

In our recordings, Session $1$ lasted \SI{60}{\second} and Session $2$ lasted \SI{40}{\second}. We used a window width of 3 seconds (\( w = \SI{3}{\second} \)) with a 2-second overlap, corresponding to a step size of 1 second \(( s = \SI{1}{\second} \)). For a session of duration \( T \), the number of generated windows \( N_w \) is computed as:
\begin{equation}
N_w = \left\lfloor \frac{T - w}{s} \right\rfloor + 1
\end{equation}
Accordingly, Session 1 yields \( N_w = \left\lfloor \frac{60 - 3}{1} \right\rfloor + 1 = 58 \) reference points per artwork, while Session 2 produces \( N_w~=~\left\lfloor \frac{40 - 3}{1} \right\rfloor + 1 = 38 \)  reference points.

We computed the mean RSS value for each beacon within each window based on all RSS samples during the window interval. This averaging helps reduce the impact of signal fluctuations and noise. Beacons that are not detected within a window are assigned a default value of $+$\SI{+110}{\decibelm}. Table~\ref{tab:fingerprint_structure} provides an example of the resulting fingerprint structure.
\begin{table}[!h]
\centering
\caption{Example of RSS fingerprints from Session 1 (Android)}
\label{tab:fingerprint_structure}
\begin{tabular}{ccccccc}
\toprule
\textbf{12-2} & \textbf{12-3} & \textbf{...} & \textbf{28-12} & \textbf{28-13} & \textbf{room} & \textbf{artwork} \\
\midrule
-97.44 & -95.11 & ... & 110.0 & 110.0 & 16 & 61 \\
-98.80 & -96.00 & ... & 110.0 & 110.0 & 16 & 61 \\
\bottomrule
\end{tabular}
\end{table}

\begin{table*}[htbp]
\centering
\caption{Performance evaluation (\%).}
\label{tab:baseline_results}
\begin{tabular}{lccccccccc}
\toprule
\multicolumn{1}{c}{} & \multicolumn{3}{c}{\textbf{Room Accuracy}} & \multicolumn{3}{c}{\textbf{Artwork Top-1 Accuracy}} & \multicolumn{3}{c}{\textbf{Artwork Top-3 Accuracy}} \\
\cmidrule(lr){2-4} \cmidrule(lr){5-7} \cmidrule(lr){8-10}
\textbf{Dataset} & \textbf{Proximity} & \textbf{1-NN} & \textbf{57-NN} & \textbf{Proximity} & \textbf{1-NN} & \textbf{57-NN} & \textbf{Proximity} & \textbf{1-NN} & \textbf{57-NN} \\
\midrule
BAR 1,2  & $92.92$ & $98.86$ & $\mathbf{98.89}$ & $64.18$ & $\mathbf{86.61}$ & $85.70$ & $88.86$ & $86.64$ & $\mathbf{97.43}$ \\
BAR 1,3  & $\mathbf{93.68}$ & $81.88$ & $74.23$ & $\mathbf{49.67}$ & $23.79$ & $20.33$ & $\mathbf{78.83}$ & $23.79$ & $47.70$ \\
BAR 1,4  & $\mathbf{93.63}$ & $90.91$ & $86.14$ & $\mathbf{44.06}$ & $33.01$ & $23.92$ & $\mathbf{75.15}$ & $33.01$ & $54.21$ \\
BAR 2,1  & $92.70$ & $99.92$ & $\mathbf{99.96}$ & $66.51$ & $88.26$ & $\mathbf{88.74}$ & $88.52$ & $88.26$ & $\mathbf{98.26}$ \\
BAR 2,3  & $\mathbf{92.36}$ & $82.01$ & $68.93$ & $\mathbf{48.08}$ & $19.20$ & $16.88$ & $\mathbf{78.98}$ & $19.20$ & $40.06$ \\
BAR 2,4  & $\mathbf{94.30}$ & $87.28$ & $80.15$ & $\mathbf{44.47}$ & $24.80$ & $23.60$ & $\mathbf{76.35}$ & $24.80$ & $50.47$ \\
BAR 3,1  & $93.79$ & $97.49$ & $\mathbf{97.89}$ & $\mathbf{60.17}$ & $46.48$ & $42.43$ & $\mathbf{89.92}$ & $46.48$ & $75.11$ \\
BAR 3,2  & $92.57$ & $98.33$ & $\mathbf{98.68}$ & $\mathbf{57.25}$ & $35.32$ & $32.43$ & $\mathbf{87.66}$ & $35.32$ & $67.28$ \\
BAR 3,4  & $93.68$ & $99.88$ & $\mathbf{99.97}$ & $43.04$ & $\mathbf{68.01}$ & $67.11$ & $78.89$ & $68.10$ & $\mathbf{96.14}$ \\
BAR 4,1  & $91.93$ & $98.49$ & $\mathbf{99.27}$ & $\mathbf{47.28}$ & $47.05$ & $46.70$ & $\mathbf{84.06}$ & $47.09$ & $78.05$ \\
BAR 4,2  & $94.21$ & $\mathbf{97.69}$ & $97.34$ & $\mathbf{50.32}$ & $39.97$ & $39.68$ & $\mathbf{87.92}$ & $39.97$ & $67.19$ \\
BAR 4,3  & $91.97$ & $97.36$ & $\mathbf{97.61}$ & $42.82$ & $\mathbf{63.20}$ & $61.02$ & $76.11$ & $63.31$ & $\mathbf{90.02}$ \\
\bottomrule
\end{tabular}
\end{table*}


\section{Baselines and Evaluation}
\label{sec:baselines}
The primary goal of the dataset is to evaluate robust position classification in complex indoor cultural heritage environments with constrained radio propagation characteristics.

We have considered 12 datasets for evaluation, all the possible combinations with the 4 subsets of data available (see Table~\ref{tab:subsets}). 
The datasets are named \emph{BAR x,y}, where $x$ indicates the subset used for training and $y$ indicates the subset used for testing.

\begin{table}[!hbt]
    \centering
    \tabcolsep 4pt
    \caption{Subsets available from the data collections}
    \label{tab:subsets}
    \begin{tabular}{cccrc}
    \toprule
         \textbf{Subset id} & \textbf{Collection} & \textbf{\# of fingerprints} & \multicolumn{2}{c}{\textbf{Valid RSS}}\\
    \midrule
         1& Android Session 1 & 5,220 & 197,789  &(40.74\%)\\ 
         2& Android Session 2 & 3,420 & 128,058 &(40.26\%)\\  
         3& IOS Session 1 & 5,220 & 74,080 &(15.26\%)\\
         4& IOS Session 2 & 3,420 & 55,288 &(17.41\%) \\
    \bottomrule
    \end{tabular}

\end{table}

For baseline purposes, we have implemented a proximity-based algorithm using the strongest RSS in the fingerprint. In the offline phase, for each beacon, we identify the fingerprints from the training set where it reports the strongest RSS. Then, we compute the probability of belonging to each room and artwork. At the operational phase, we use the previously computer  computed probabilities and the strongest RSS in the operational fingerprint to estimate the artwork and room. We also implement the $k$-Nearest Neighbors ($k$-NN) classifier due to its computational efficiency and signal-space mapping capabilities. We parameterize the classifier with two distinct configurations: $k=1$ (1-NN), representing the minimal implementation with lowest computational overhead, and $k=57$ to leverage more samples from the training set. In fact, larger $k$ values generally provide superior classification performance in environments with high signal fluctuation, effectively creating a robust majority voting mechanism that reduces the impact of outlier measurements commonly found in spaces with complex radio propagation characteristics. Since the dataset was recorded during opening hours, the RSS signal may be affected by multipath effects.

Table~\ref{tab:baseline_results} presents our benchmark results. To evaluate positioning performance systematically, we propose three complementary metrics:
\begin{itemize}
    \item \textbf{Room Accuracy}: Quantifies the system's capability to correctly identify the spatial compartment in which a user is located, establishing zone localization performance.
    
    \item \textbf{Artwork Top-1 Accuracy}: Measures the precision with which the algorithm correctly identifies the nearest POI, based on signal strength as its highest probability prediction.
    
    \item \textbf{Artwork Top-3 Accuracy}: Evaluates positioning robustness by considering success if the correct POI appears among the algorithm's three highest-probability predictions, accounting for signal multipath effects.
\end{itemize}

In the context of cultural heritage facilities, high room accuracy enables efficient spatial monitoring for occupancy analysis and visitor flow optimization. The top-1 accuracy metric for artwork proximity detection quantifies the hit rate of the actual nearest POI. Acknowledging that BLE-based systems typically exhibit localization errors in the order of \SIrange{2}{3}{\meter} due to signal propagation characteristics and multipath interference~\cite{torres23}, we therefore implement the top-3 accuracy criterion to compensate for these inherent radio frequency propagation limitations. This relaxed constraint is particularly valuable for interactive applications that can present multiple content options to users who can then select the appropriate exhibit based on visual confirmation.

These metrics comprehensively characterize the system performance across multiple dimensions relevant to indoor positioning applications, where utility depends on both spatial precision and contextual awareness at different granularities. Moreover, the proposed evaluation methodology systematically assesses positioning performance across three distinct signal processing scenarios: cross-session (temporal variance), cross-platform (hardware heterogeneity), and combined cross-session and cross-platform testing (comprehensive real-world deployment conditions).

\section{Dataset Use Cases}

\subsection{Cross-Platform Analysis}
Numerous studies have explored the differences in device antennas, but our dataset enables us to study a new class of devices that is underrepresented in the available datasets. The dataset offers simultaneous recordings from iOS and Android devices, allowing investigations on platform-specific characteristics significantly impacting positioning accuracy. 

Our analysis reveals considerable challenges in accurately predicting user positions when the training and testing sets are recorded with different devices, even within the same session. For instance, the $57$-NN Artwork Top-3 Accuracy is between \SIrange{40.06}{54.21}{\percent} when using the Android sessions for training and the IOS sessions for testing (datasets \emph{BAR 1,3}; \emph{BAR 1,4}; \emph{BAR 2,3}; and \emph{BAR 2,4}), while it is between \SIrange{67.19}{78.05}{\percent} when using the IOS sessions for training and the Android sessions for testing (datasets \emph{BAR 3,1}; \emph{BAR 3,2}; \emph{BAR 4,1}; and \emph{BAR 4,1}). 
Conversely, the $57$-NN Artwork Top-3 Accuracy is much higher when using the same platform for training and testing with values between \SIrange{97.43}{98.26}{\percent} when using solely the Android sessions (datasets \emph{BAR 1,2}; \emph{BAR 2,1}), and values between \SIrange{90.02}{96.14}{\percent} when using solely the IOS sessions (datasets \emph{BAR 3,4}; \emph{BAR 4,3}).

\begin{figure}[h]
    \centering
    \includegraphics[width=0.95\linewidth]{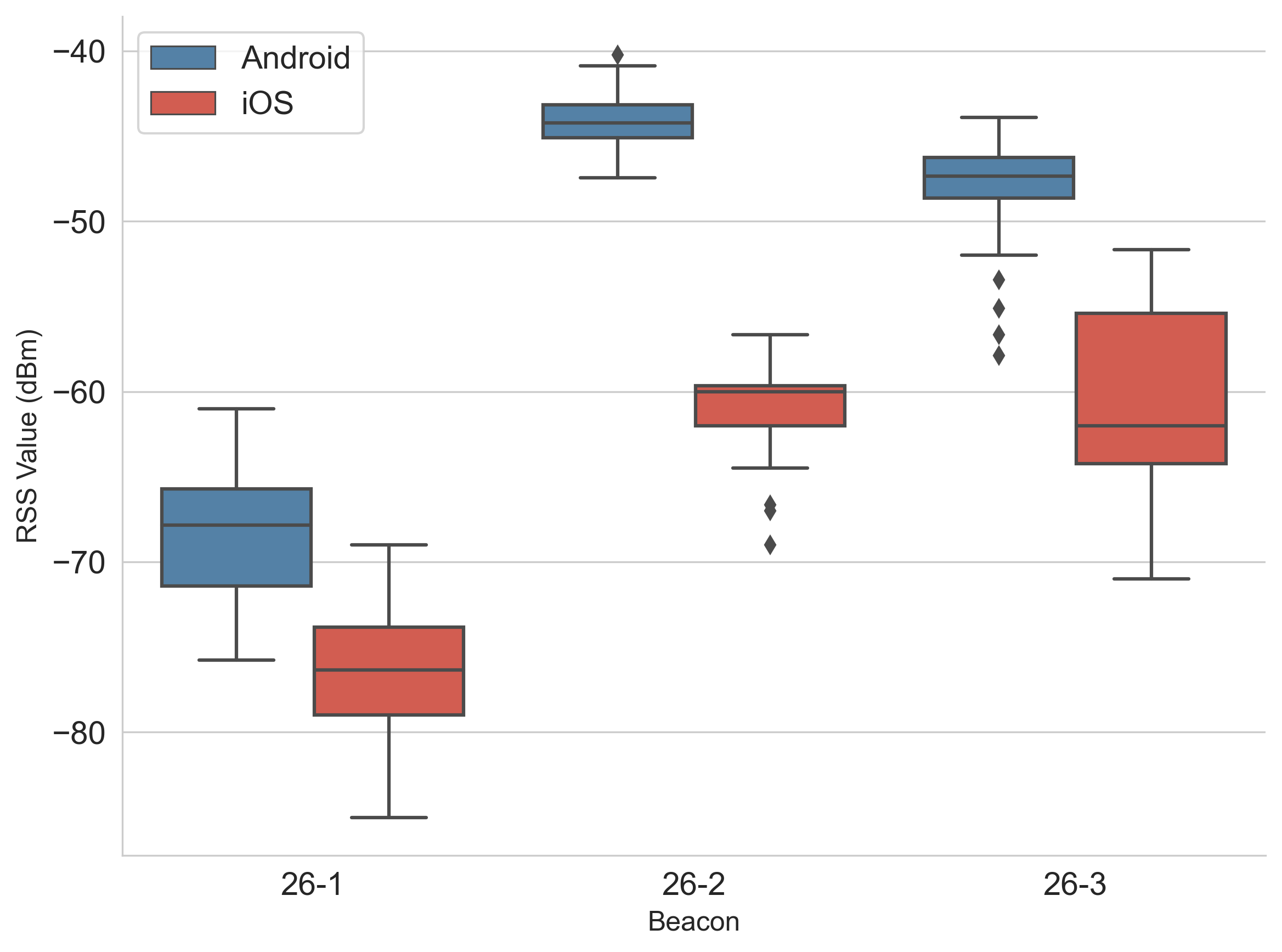}
    \caption{Comparison of RSSs between iOS and Android devices in front of the same artwork during the same session. Box plots show median, quartiles, and outliers for each platform.}
    \label{fig:platform_comparison}
\end{figure}

Fig.~\ref{fig:platform_comparison} illustrates the distribution of RSSs for the same beacons measured simultaneously across platforms. The exploitation of such data may have important implications for cross-platform deployment scenarios. The differences in antenna design between platforms lead to consistently different RSS measurements, creating a fundamental challenge that must be addressed for reliable cross-device positioning. 

The platform-specific signal variations could be managed through calibration that normalizes readings between devices. By applying appropriate correction factors based on antenna features, researchers can develop positioning methods, maintaining consistent performance regardless of the user's device.

\subsection{Anomaly Detection and System Resilience Studies}
\begin{figure}[htbp]
    \centering
    \includegraphics[width=1\linewidth]{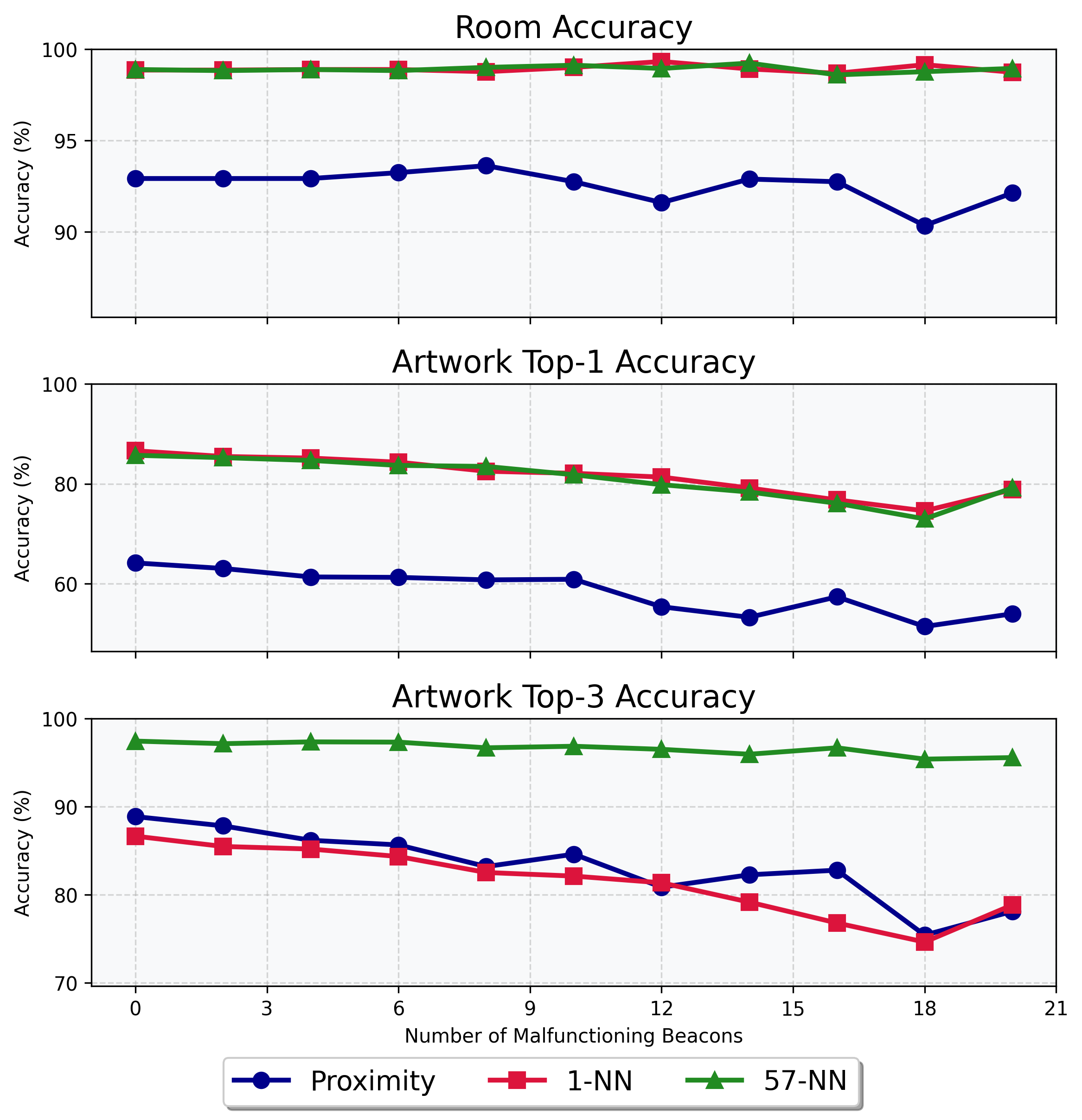}
    \caption{Results of positioning simulating the malfunction of beacons.}
    \label{fig:anomalies}
\end{figure}

Another research opportunity is related to the automated detection of system anomalies, such as beacon failures or displacements. For example, we conducted experiments to quantify the impact of malfunctioning beacons on positioning accuracy by systematically removing RSS data from beacons (i.e., setting columns to non-detected) during operational phase using the combination \emph{BAR 1,2} (Fig.~\ref{fig:anomalies}).

The results reveal distinct resilience patterns across the performance metrics. Room Accuracy remains remarkably stable even as beacon failures increase. Artwork Top-1 Accuracy demonstrates a clear downward trend for all methods as beacon failures increase. Finally, Artwork Top-3 Accuracy shows significant algorithm-dependent resilience differences. The 57-NN approach maintains stability while both 1-NN and Proximity methods show noticeable degradation patterns, declining from approximately $85-89\%$ to around $75\%$ accuracy, with some fluctuations, as failures increase.

These findings highlight two important observations: first, room-level positioning remains robust even with substantial infrastructure failures; second, artwork-level accuracy (especially Top-1) is more vulnerable to beacon malfunctions. The superior resilience of the 57-NN algorithm across all metrics suggests that more sophisticated neighbor-based approaches can mitigate the impact of system anomalies.
The dataset enables the development and testing anomaly detection techniques that could automatically identify when beacons malfunction or are moved from their original positions.

\subsection{System Deployment Considerations}
Practical deployment of positioning systems in cultural heritage spaces involves balancing technical performance against environmental constraints. Fig.~\ref{fig:anomalies} in addition to demonstrating the accuracy degradation, shows also some interesting fluctuations in accuracy when a particular set of beacons is removed. For instance, at certain points (such as between 15-18 malfunctioning beacons), we observe brief performance recoveries or plateaus across multiple algorithms, suggesting that not all beacons may contribute equally to positioning accuracy. This observation opens up opportunities for research into beacon placement optimization. By analyzing which beacons, when removed, cause minimal performance degradation, we can identify a minimal effective subset of transmitters that maintains acceptable accuracy levels. Additionally, the resilience data suggests that different positioning tasks (room-level vs. artwork-level) have varying sensitivity to beacon reduction. This could lead to hierarchic deployment strategies where critical areas receive denser beacon coverage while less demanding zones operate with minimal infrastructure. 

\section{Conclusions}
\label{sec:discussionConclusion}
In this paper, we presented and analyzed \emph{BAR}, a novel BLE RSS dataset collected in a real-world museum environment. Our dataset addresses significant gaps in publicly available datasets for cultural heritage spaces by providing comprehensive coverage of $90$ artworks across $13$ rooms, cross-platform measurements from both iOS and Android devices, and advanced baselines for performance evaluation. The results show interesting patterns when it comes to cross-platform evaluation and further demonstrate that IPSs can be leveraged to get visitor position with high accuracies.

However, this study comes with certain limitations. Specifically, all recordings were made by a single person using only one device for each platform, thus limiting the data diversity. Additionally, we applied a specific proximity algorithm, therefore it is possible that different implementations could yield even better results. Finally, we employed just one technology, even though particularly widespread in indoor positioning for its satisfactory balance between cost and benefits.

Following the principles highlighted in~\cite{ANAGNOSTOPOULOS2025101485}, we provide detailed documentation of our environment setup, collection methodology, and data structure to ensure reproducibility and facilitate future research. Moreover, we make all the code publicly available~\cite{maestro2025flutter,ferrato25}.

Finally, we have shown how our dataset enables different research directions including cross-platform studies, system resilience exploration, and optimization of beacon deployment. These opportunities can drive innovation in IPSs specifically tailored to the unique constraints of cultural heritage environments. We plan to expand the dataset with additional collection sessions using different devices, explore sensor fusion techniques that leverage the recorded accelerometer, gyroscope, and magnetometer data, and develop adaptive algorithms that can maintain high positioning accuracy while reducing the computational cost.

\section*{Acknowledgment}
The authors wish to express sincere gratitude to the \textit{Gallerie Nazionali di Arte Antica}, and especially to Prof. Dr. Michele Di Monte and Dr. Paola Guarnera, for granting permission to conduct and promote this research. J. Torres-Sospedra acknowledges funding from Generalitat Valenciana (CIDEXG/2023/17, Conselleria d’Educació, Universitats i Ocupació).

\section*{Data Usage and Reproducibility}

The new datasets presented in this work are publicly available in \cite{ferrato25}. Additional supplementary materials supporting research reproducibility are publicly available in \cite{maestro2025flutter,ferrato25}.



\end{document}